\documentclass[10pt, conference, compsocconf]{IEEEtran}
\ifCLASSINFOpdf
  \usepackage[pdftex]{graphicx}
\else
\fi
\usepackage[utf8]{inputenc}

\begin{document}
\title{Digitizing Handwriting with a Sensor Pen: \\
A Writer-Independent Recognizer}


\author{\IEEEauthorblockN{Mohamad Wehbi}
\IEEEauthorblockA{Machine Learning and Data Analytics Lab\\
Friedrich-Alexander-Universität Erlangen-Nürnberg\\
Erlangen, Germany\\
mohamad.wehbi@fau.de}
\and
\IEEEauthorblockN{Tim Hamann}
\IEEEauthorblockA{STABILO International GmbH\\
%line 2: name of organization, acronyms acceptable\\
Heroldsberg, Germany\\
tim.hamann@stabilo.com}
\and
\IEEEauthorblockN{Jens Barth}
\IEEEauthorblockA{STABILO International GmbH\\
%line 2: name of organization, acronyms acceptable\\
Heroldsberg, Germany\\
jens.barth@stabilo.com}
\and
\IEEEauthorblockN{Bjoern Eskofier}
\IEEEauthorblockA{Machine Learning and Data Analytics Lab\\
Friedrich-Alexander-Universität Erlangen-Nürnberg\\
Erlangen, Germany\\
bjoern.eskofier@fau.de}
}


%Telephone: (800) 555--1212, Fax: (888) 555--1212}


\maketitle

\begin{abstract}
Online handwriting recognition has been studied for a long time with only few practicable results when writing on normal paper. Previous approaches using sensor-based devices encountered problems that limited the usage of the developed systems in real-world applications. This paper presents a writer-independent system that recognizes characters written on plain paper with the use of a sensor-equipped pen. This system is applicable in real-world applications and requires no user-specific training for recognition. The pen provides linear acceleration, angular velocity, magnetic field, and force applied by the user, and acts as a digitizer that transforms the analogue signals of the sensors into timeseries data while writing on regular paper. The dataset we collected with this pen consists of Latin lower-case and upper-case alphabets. We present the results of a convolutional neural network model for letter classification and show that this approach is practical and achieves promising results for writer-independent character recognition. This work aims at providing a real-time handwriting recognition system to be used for writing on normal paper.

\end{abstract}


% \fi
\IEEEpeerreviewmaketitle

\section{Introduction}

Handwriting recognition (HWR) is the process of converting one's handwriting into a format that a computer recognizes. The field of HWR has gained a renewed interest over the past years due to the high demand of recognition applications for improved human-computer interaction. HWR is classified mainly into two types, offline and online recognition \cite{Priya2016OnlineAO}.

\begin{figure}[!t]
\centering
\includegraphics[width=0.4\textwidth]{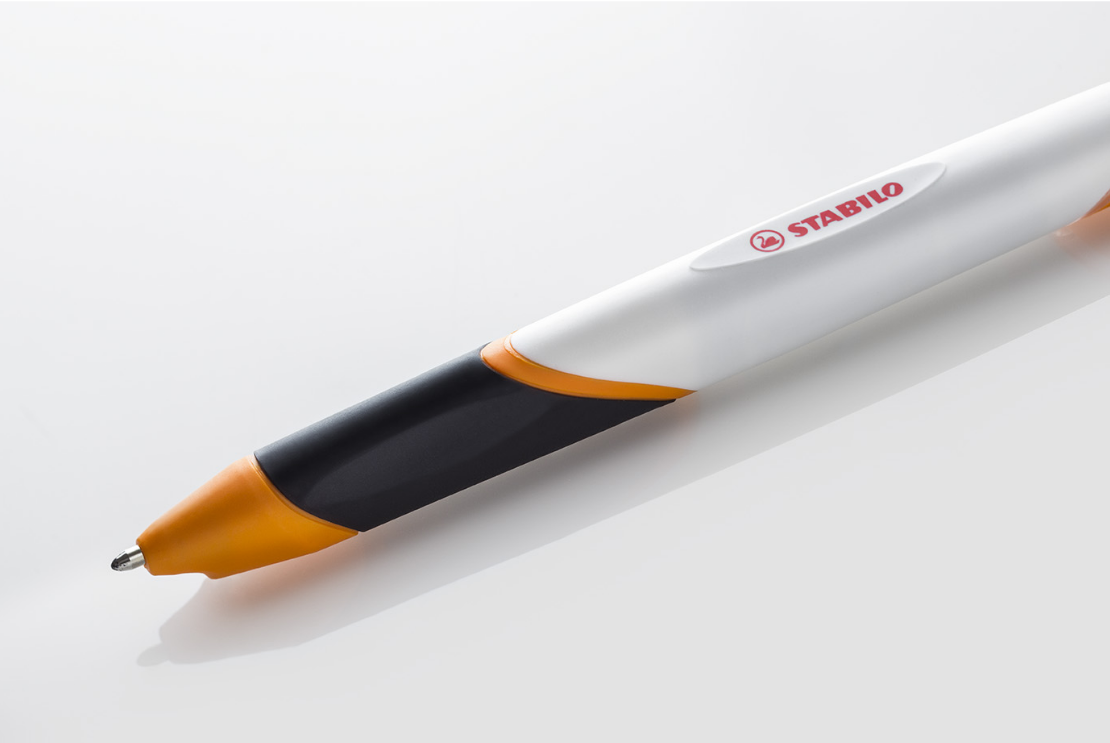}
\caption{The Digipen by STABILO \cite{b14}.}
\label{fig1}
\end{figure}

Offline HWR, also known as optical character recognition (OCR), takes place when only static information is available to the system. It is applied after a document is finished by the writer with the use of cameras or scanners that convert the handwriting into a digital form. This has been implemented in different application scenarios such as recognizing numbers on bank checks\cite{Palacios2004HandwrittenBC}, understanding addresses on mail\cite{Srihari1993RecognitionOH}, and reading forms\cite{Garris1998IntelligentSF}. Offline HWR can be useful in different domains and can be applied at any time after a document is written. However, a disadvantage of offline HWR is that it cannot be used when trying to apply recognition in real-time as a person is writing.

%\begin{figure}[]
%\centerline{\includegraphics[width=0.5\textwidth]{imgs/fig1.png}}
%\caption{The Digipen by STABILO \cite{b14}.}
%\label{fig1}
%\end{figure}

Online HWR (OHWR) is the analysis of spatiotemporal signals with the purpose of understanding a geometrical design. An online recognizer understands the dynamic movement of the writer's hand, which includes the writing speed and direction. It could also include different applied strokes while writing. Online HWR systems can be more accurate than offline systems since temporal information is also passed along with the spatial information, as opposed to offline systems where only static information is available. Spatiotemporal information can be helpful when trying to distinguish between similarly shaped characters by knowing the number of strokes that were needed for each character, and following different movement pauses that occur during writing when characters contain corners. In other cases, this dynamic information can confuse the system when dealing with characters that can be written in multiple numbers of strokes which depends on the writer. In English, the average number of strokes is two strokes for uppercase letters and one stroke for lowercase letters. Cursive writing has an average of less than one stroke per letter\cite{Tappert2007CHAPTER6}.

Applications of OHWR systems vary across many domains. An OHWR system is useful for different users since it allows them to obtain the advantages of writing, with saving time by avoiding the scanning documents for digitization. Some applications are handwriting analysis in children\cite{PazVillagrn2014LiftsAS}, detecting alcoholism\cite{Shin2014DetectionOA}, and neurological analysis of patients\cite{Dounskaia2009BiasedWA}.

For an OHWR system, a digitizer is required for the digitization process. Handheld devices such as tablets and smartphones recognize handwriting by following the location of the finger or stylus pen tip on the screen of the device. This has been successfully implemented in different applications but is constrained to the use of special writing surfaces. Gerth et al. \cite{Gerth2016IsHP} have shown that there are differences between writing on paper and writing on a tablet surface, where the latter requires additional movement control and presents a more challenging writing process. Therefore, to make recognition possible while writing on regular paper surfaces, devices with different sets of inertial measurement unit (IMU) sensors are required to track the movement of the writer's hand. This has been implemented in several applications that are discussed in the next section, yet the results are still not satisfactory to allow such systems to be used in real-world applications for real-time recognition systems.

When applying an OHWR system for real-world usage, different aspects should be considered for the system to be usable. One main requirement is that the system is simple to use, i.e no complex hardware with complex configuration needed. Another important aspect is that the system should work for all writers, that is, recognition should be writer-independent, and not trained for a specific writer. In this paper, we present a  writer-independent system trained to recognize the Latin alphabet with the use of a sensor-enhanced ballpoint pen presented in Fig.\,\ref{fig1}. The recognizer requires no prior knowledge of the writer and can be implemented on a tablet for real-time recognition while writing on plain paper.

\section{Related Work}

Over the past few years, several sensor-equipped devices have been designed and different approaches have been used to solve the task of real-time handwriting recognition.

In combining OCR with an IMU-based pen, Wang et al. \cite{Wang2010AnIP} pursued the idea of using a trajectory reconstruction algorithm to estimate the position of the tip of a pen. The resulting trajectories were then classified by a regular optical digit recognition software on a tablet computer. They achieved a recognition rate of 94.6\% for the ten digits.

\begin{figure}[!t]
\centerline{\includegraphics[width=0.4\textwidth]{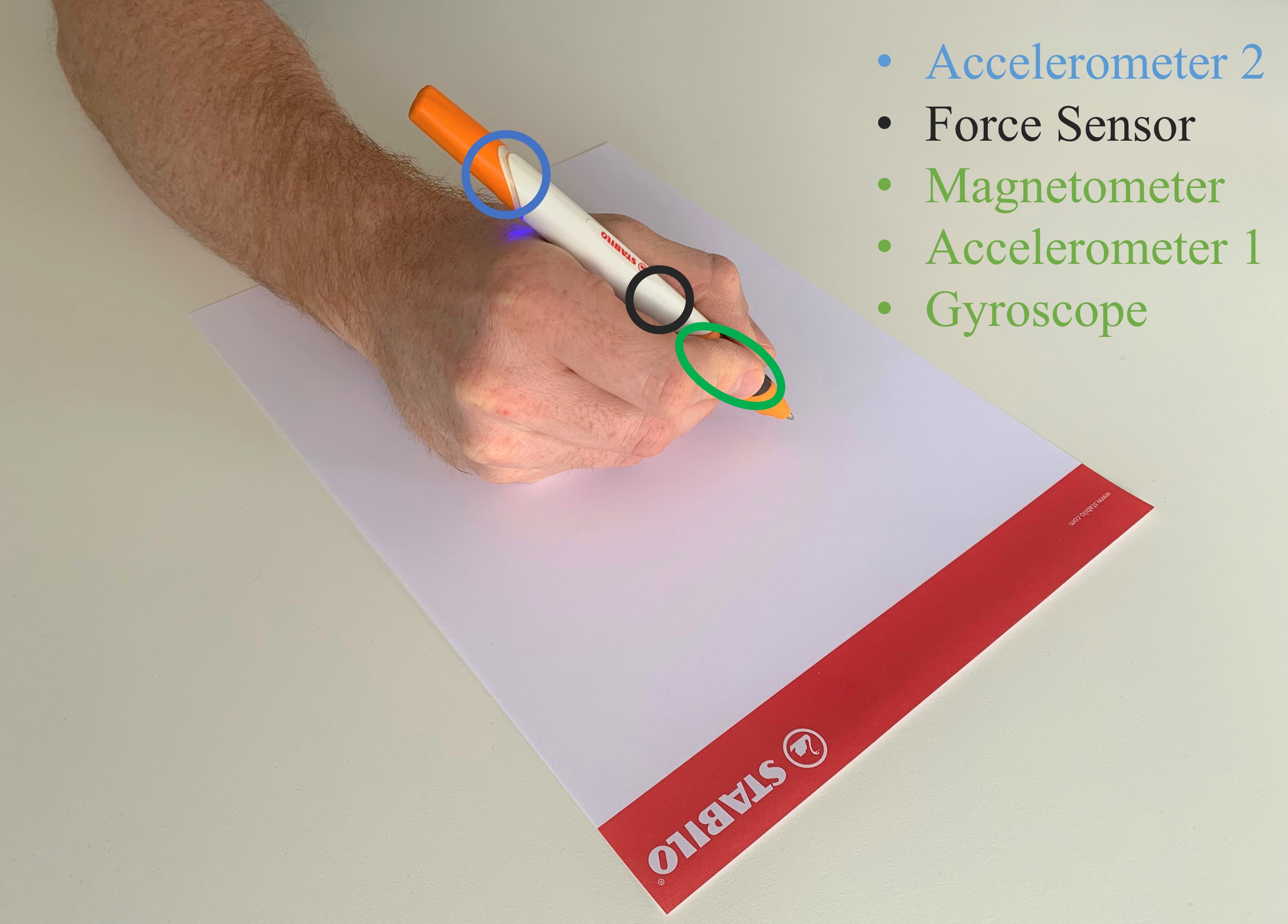}}
\caption{The location of the sensors on the Digipen \cite{b14}.}
\label{fig6}
\end{figure}

Deselaers et al. \cite{Deselaers2015GyroPenGF} used gyroscope data from a mobile phone to trace an image of the hand movement while writing and used the produced image to recognize the written word following known offline writing recognition methods. They applied the system for single stroke movement of the device on the writing surface and presented the results showing the tracing accuracy of the hand movement.

Amma et al. \cite{Amma2010AirwritingRU} used a data glove to measure hand motion and extract gestures and characters drawn into the air. The tool was equipped with gyroscopes and accelerometers. They achieved a writer-dependent character recognition rate of 94.8\% and a writer-independent rate of 81.9\% using Hidden Markov Models. However, the complexity of the hardware with the process of writing in air posed a limitation for the usage of such a system as a recognizer application that can be used in regular daily situations.

Schrapel et al. \cite{Schrapel2018PentelligenceCP} presented a sensor-equipped pen called Pentelligence. It senses the pen tip's motions and records the sound emissions when performing strokes. These data were used to train a Neural Network. They achieved accuracies of 78\% (writer-independent) and 98\% (writer-dependent) when classifying the ten digits.

Koellner et al. \cite{Koellner2019WhatDY} used the Digipen by STABILO\footnote{STABILO International GmbH (Heroldsberg, Germany) } to further pursue the research of online character recognition over 26 lowercase Latin letters with a collected dataset of 20,000 samples. A comparison is presented between LSTMs and other machine learning models for OHWR achieving 94\% accuracy for writer-dependent classification, while showing a 52\% accuracy for writer-independent classification, making the system inapplicable to be used for general recognition.

Summing up, different approaches have been taken to solve the problem of OHWR with different devices. However, various restrictions such as recognizing digits only, being user-dependent, or covering single stroke writing, and even using complex hardware limit the success of such systems in real-world applications. With such limitations, commercial HWR applications tend to apply OCR methods on images produced by tracing a pen trajectory over a specific writing surface. Thus, we present a system using a sensor pen input device for online character recognition which, in contrast to the related work, relies on a pen without the need for additional hardware, and based on the research done, we provide better results over the Latin alphabet for writer-independent recognition.

\section{Methodology}

\subsection{Digipen}

The approach presented in this paper uses the STABILO Digipen, which is a sensor-enhanced ballpoint pen equipped with five sensors: Two accelerometers, a gyroscope, a magnetometer, and a force sensor distributed over the pen as shown in Fig.\,\ref{fig6}. The accelerometers were adjusted to a range of $\pm$\,2g with a resolution of 16 Bit (front) and 14 Bit (rear), the gyroscope has a range of $\pm$\,1000\,$^{\circ}$/s with a resolution of 16 Bit, the magnetometer has a range of 2.4\,mT with a resolution of 14 Bit and the force sensor has a measurement range of 0 to 5.32\,N with a resolution of 12 Bit. The pen is also provided with a soft-touch grip zone for holding the pen, and includes a Bluetooth module that enables the live streaming of data to a handheld tablet or smart phone. The data provided by the recordings were used in its raw form without the need for further calibration.

\subsection{Data Acquisition}

The approach presented in this paper utilizes the pen's raw sensor data to classify letters. For data collection, a mobile app was provided by STABILO for saving the data stream while using the sensor-equipped Digipen as the input device. The pen is connected to the app via Bluetooth Low Energy and the sensor data are transmitted at a rate of 100 Hz throughout the session.

\begin{figure}[!t]
\centerline{\includegraphics[width=0.4\textwidth]{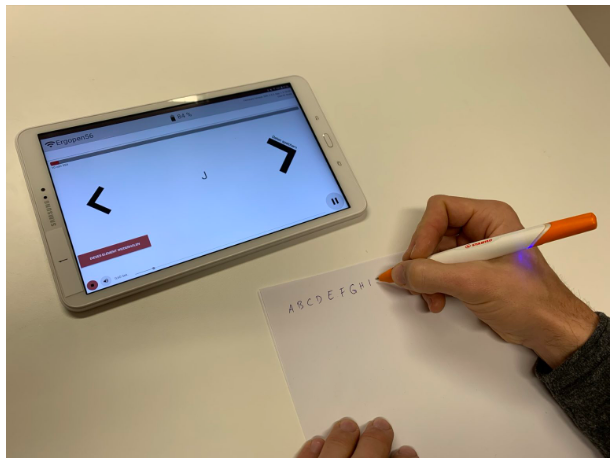}}
\caption{The data recording app provided by STABILO \cite{b14}.}
\label{fig2}
\end{figure}

To carry out the data collection process, several data recording sessions were conducted with 114 right-handed subjects who volunteered to participate. The subjects were requested to insert data such as name, age, gender, and dominant hand before the recording starts. The app displays the letter to write, as displayed in Fig. \ref{fig2}, going through the uppercase then lowercase alphabet and repeats this process up to 6 times. At the end of the recording, two files were saved on the tablet, the data stream file of all the sensor values throughout the complete recording session, and the labels file in which the ground truths are identified based on the time when they were displayed on the screen. These recordings resulted in approximately 14,870 uppercase and 14,841 lowercase letter recordings since not all subjects fully completed the recordings.

For the data acquisition process, some boundary conditions had to be met: Only right-handed persons over the age of 16 were allowed to participate in the recordings. The volunteers were instructed to hold the pen in its correct orientation (with the logo facing upward). Furthermore, the writing surface had to be horizontal and padded by at least 5 paper sheets to create homogeneous conditions throughout different recording sessions.

\subsection{Data Preparation}

Following the data collection process, the labels file of each recording session was used to split the sensor data file into single letters with the relative timeseries data. Each split single-letter file consists of the sensor data that were transmitted while the letter was shown in the recording app. This includes phases where the pen was hovering before and after writing the letter.

\begin{figure}[!t]
\centerline{\includegraphics[width=0.4\textwidth]{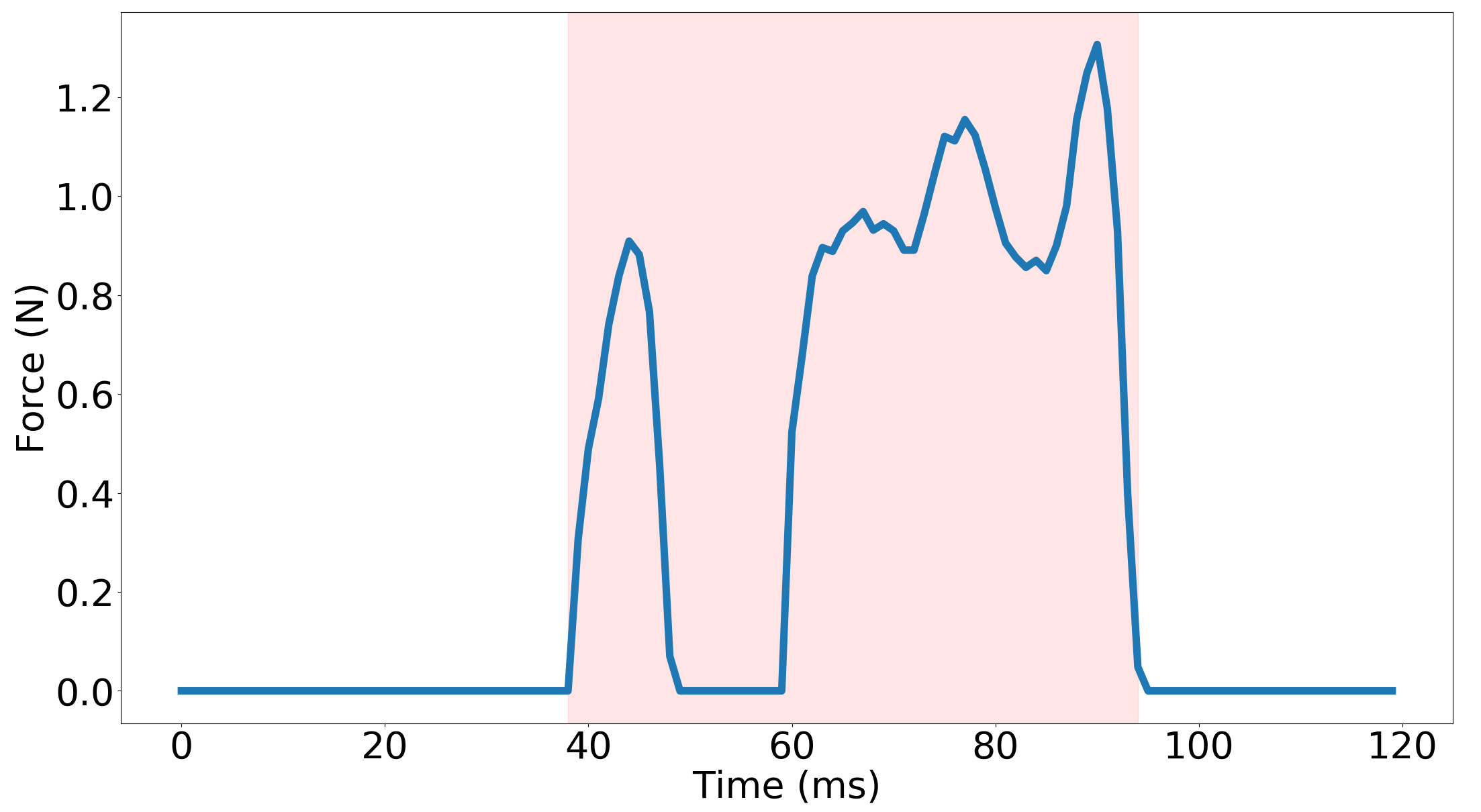}}
\caption{A graph showing the raw force sensor values of a single letter with the highlighted part of the recording to be used as a single input sample for the neural network.}
\label{fig3}
\end{figure}

To train only on meaningful data, each sample was filtered keeping only the timesteps when the user was writing and not hovering between two different letters. This was applied by removing all datapoints of a sample with a force sensor value less than 0.2~N, before and after writing the letter, while keeping the low values in between the start and end of a letter, to represent multiple strokes. Fig. \ref{fig3} shows a force sensor recording of the letter 'B', where the first peak in the signal represents the writing of the vertical line, and the other peaks representing the writing of its two semi-circles.

However, removing the datapoints with low force signals resulted in the samples having different lengths, which was mainly due to different writing speeds between different subjects, or between different letters of a single subject. Thus a resampling technique was required to equalize the length of the samples.

Each letter sample was linearly interpolated to a length of 50 timesteps, which represented the mean length of all the data samples collected. The interpolation allowed an easier fitting of the data into the model and was assumed to regularize the pace of writing that helped the model to better learn and classify the data. Samples with a length outside the interval of 10-500 timesteps were considered as outliers and removed from the dataset. Finally, each sample was normalized using L2 feature-wise normalization which allowed the model learning process to run more smoothly and provided better results.

\subsection{Network Architecture \& Training}

Previous works with similar data types mainly used a Recurrent Neural Network (RNN) model since RNNs were designed to work with sequential data to retain and remember information throughout a long sequence, and are also easier to use when dealing with data of unknown lengths. However, LSTMs performed poorly when dealing with writer-independent recognition, which is mainly because a sequence of the same letter changes from one writer to another. 

The model used in this paper was based on a 1D-Convolutional Neural Network (CNN). The data samples were of fixed lengths after the interpolation, which was necessary to fit the data into a CNN. The usage of a convolutional model was due to the ability of CNNs to capture local dependencies and scale-invariant features of signals\cite{Zeng2014ConvolutionalNN}. The convolutional model was able to extract important features without the need for any further preprocessing of the data.

\begin{figure}[]
\centerline{\includegraphics[width=.4\textwidth]{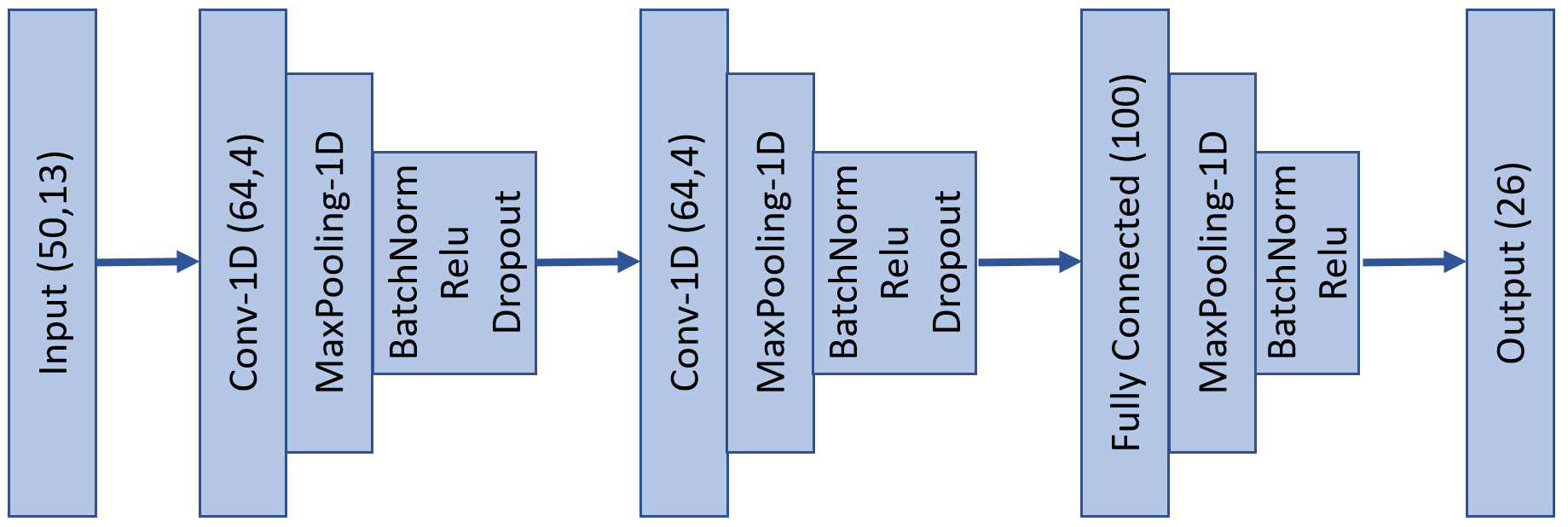}}
\caption{A summary of the network architecture used.}
\label{fig4}
\end{figure}

The model was implemented with the Keras library\cite{Chollet2018KerasTP}. It consisted of a 13 channel input layer (12 channels for the x,y,z axis of each of the four tri-axial sensors and one channel for the force sensor), followed by three hidden layers, with an output layer of 26 classes. The training of the network was done over 50 epochs. The 'Relu' activation function was used for all hidden layers, with a 'Softmax' activation at the output layer. BatchNormalization was done after all hidden layers, and dropout was applied at the first and second hidden layers with a 40\% dropout rate. As a loss function, the 'categorical crossentropy' loss was used, and 'Adam' optimizer with a learning rate of 0.001. The convolutional layers consisted of 64 filters each, with a kernel size of 4, and the fully connected hidden layer consisted of 100 units. The same model configuration was used for both uppercase and lowercase letter classification. Fig. \ref{fig4} shows a summary of the architecture used in the model.

To better provide the reader with a comparison to previous works, another model was implemented with an LSTM architecture with two hidden layers using the 'Relu' activation function. The other hyperparameters of the network and the training were left unchanged. This was done to show how the different parts of our experiments contributed to providing a better recognition accuracy.

\section{Results}

We show the results obtained by using the data collected with the STABILO Digipen by the CNN and the LSTM models. This work focuses primarily on the writer-independent character recognition since writer-dependent is in most cases not applicable for a general real-world use-case scenario. However, since in some commercial application scenarios, it is common to have the user adapt the pre-trained model with some individual training data to enable personalization, we run experiments over a writer-dependent recognition scenario and include the results.

The complete dataset was used in both scenarios. For the writer-independent case, we exclude a number of subjects from the training dataset for testing, while in the writer-dependent case, a writer that contributed to the training set can also contribute samples to the test set.

The models were trained for the uppercase and lowercase classification separately. The accuracy was the metric used for the evaluation. Since the writing shape variance for the uppercase letters is lower than that in the lowercase letters, the accuracy was assumed to be higher in the former case. The results of each of the two scenarios are discussed separately.

\subsection{Writer-dependent case}

A system that has seen samples of how a specific user writes is suspected to find it easier to recognize other samples of this specific user, since each individual has a certain way of holding the pen, moving the pen, writing letters in a specific way, and specific writing speed. So, having learned beforehand how a writer writes, a model can use these qualities for better recognition of what is written. Given also how that dataset was recorded, having randomly split the data into training and test sets, a specific sample cannot be in both sets, but it is likely that another sample of the same person would be available in the test set since every subject recorded the complete alphabet more than once.

%\begin{figure*}[!t]
%\centerline{\includegraphics[width=1\textwidth]{imgs/fig7.png}}
%\caption{Uppercase letters confusion matrix of the CNN model for writer-independent classification presented in percentage.}
%\label{fig7}
%\end{figure*}

For this experiment, the data is split randomly by a ratio of 80/20 between training and test sets. This was done over 5 repetitions with shifting train and test sets using a 5-fold cross-validation method. The final accuracy was calculated by averaging the accuracies over the 5 runs. 

Applying the CNN model over the uppercase dataset yielded a mean accuracy of 91.04\% successful recognitions with a standard deviation of $\pm$\,0.34\%, while the LSTM model resulted in an accuracy of 84.83\% $\pm$\,0.27\%.
The same models were applied over the lowercase dataset separately and yielded a mean accuracy of 86.27\%  with a standard deviation of $\pm$\,0.75\% for the CNN and 72.01\% $\pm$\,0.87\% for the LSTM.

\subsection{Writer-independent case}

We present the results of the main focus of this paper. This case describes a system that is able to recognize input samples from subjects whose handwriting has not been seen before, simulating a completely writer-independent environment relying solely on the training data given from different subjects who contributed to the training set.

To train the model, the subjects were divided, leaving 24 subjects to be used as a test set with training done over the other 90 subjects. As in the writer-dependent case, the experiment was done over 5 runs with different train and test sets, and the final accuracy was then calculated based on the mean of the different accuracies.

Similarly as before, we run the models over the uppercase and lowercase datasets separately. The accuracy obtained using the CNN model was 86.97\% $\pm$\,1.19\% and 81.13\% $\pm$\, 2.52\% standard deviation, whereas the accuracies obtained using the LSTM model were 81.45\% $\pm$\, 1.46\% and 70.04\% $\pm$\, 2.79\% respectively. %A confusion matrix for the 26 uppercase letters is presented in Fig. \ref{fig7} that shows the predictions over the test set for a single fold in the training process. 

\section{Discussion \& Conclusion}\label{DC}

In this paper, we presented a writer-independent OHWR system that uses a sensor-equipped ballpoint pen developed at STABILO. The data used was the Latin alphabet in lowercase and uppercase letters separately. We implemented CNN and LSTM models that utilized the raw data stream from the pen which was used for the classification of 26 classes in both alphabet cases. We presented the results of running the model for two cases, the writer-dependent and the writer-independent recognition.

Considering the writer-dependent experiment results, a better recognition model with higher accuracy has been reported as presented in the related work section. However, this was not the main focus and objective of this work since this is not best applicable in a real-world use-case scenario. These results were presented for the sake of completeness of the experiment. 

The main contribution of this paper is presenting a working system for writer-independent online character recognition. Considering the related work discussed in Section II, the results presented cannot be directly comparable to most of the previous works, since the systems used are different with distinct datasets generated for the use within the experiments. However, since some of these experiments recognized digits only, or required the use of complex hardware, this indicates that our system is on a competitive level with such systems, recognizing the alphabet with using simple hardware.

A direct comparison can be done with \cite{Koellner2019WhatDY}. The Ergopen, which consists of the same hardware as the DigiPen, was used for collecting data from right-handed subjects and conducting the experiments. In the stated experiments, LSTMs outperformed other models, yet still provided low accuracy when recognizing lowercase letters from unseen subjects. The most common issue to consider with online handwriting recognition is the ability to generalize the model for subject-independent recognition which can be seen by the wide variance of results between writer-dependent and writer-independent accuracies in previous works. 

The data preparation was essential to achieve better results. The removal of uninformative timesteps from the data allowed the network to better learn features that achieved better recognition accuracies. Additionally, the interpolation of the data was also necessary to allow the network to deal with different writing speeds and paces, from unseen subjects. This can be inferred by the results of the LSTM model, which were significantly better in the writer-independent recognition in comparison with \cite{Koellner2019WhatDY}. Furthermore, the CNN architecture achieved even higher accuracies for letter recognition with faster training time in comparison to the LSTM model.

This system mostly confuses between different letters written in a similar way, like the letters 'P' and 'D', where the writer usually starts with a vertical line, followed by a semi-circle. Similarly, a high confusion rate occured in the case of the letter 'X' which is, in some cases, predicted as a 'Y' due to the shape similarity. However, considering different letters with different shapes, the model presented robust predictions with high accuracies over data from unseen subjects. It shows that a CNN model might be able to overcome the issue of generalization by understanding the different strokes of letters, creating a broader view of the sample to be recognized. However, this requires further investigation which is beyond the scope of this paper. 

This system relies solely on the pen for data transmission with a mobile app in which the model can be integrated for real-time recognition, and has been implemented in a working demo for live letter recognition also developed by STABILO upon finishing the experiments. The dataset used for this paper is intended to be released to be available for public usage in the scientific community \cite{b14}.

Future work following this will be to adjust the model for character recognition of the complete Latin alphabet with both uppercase and lowercase letters instead of separate classification. To further improve the system, the recognition should also be applicable for left-handed writers, as this system was applied for right-handed writers only since the majority of writers are of the latter case. Sahu et al. \cite{Sahu2017DifferentiationAC} have shown that there exists a significant difference between left-handed and right-handed writing which implies that the data slightly differ between the two cases. A more sophisticated system would be implemented for intra-word segmentation, then having the segmented characters classified by the letter recognition model.

\section*{Acknowledgment}

The authors would like to thank STABILO for providing what was needed for the data acquisition and the fulfillment of this work, and also the individuals from the Fraunhofer IIS, Saarland University and Kinemic GmbH who contributed in the data recording. 

We also would like to thank the Bayerisches Staatsministerium für Wirtschaft, Landesentwicklung und Energie for the funding of this work which is part of the EINNS project (Entwicklung Intelligenter Neuronaler Netze zur Schrifterkennung).

Bjoern Eskofier gratefully acknowledges the support of the German Research Foundation (DFG) within the framework of the Heisenberg professorship programme (grant number ES 434/8-1).



%\bibliographystyle{IEEEtran}

\end{document}